\pdfoutput=1

\documentclass[11pt]{article}

\usepackage{EACL2023}

\usepackage{times}
\usepackage{latexsym}

\usepackage[T1]{fontenc}

\usepackage[utf8]{inputenc}

\usepackage{microtype}

\usepackage{inconsolata}

\usepackage{graphicx}

\usepackage{multirow}
\usepackage{array}

\def\MLdel#1{\bgroup\markoverwith{\textcolor{brown}{\rule[0.5ex]{2pt}{1pt}}}\ULon{#1}}

\def\MLdel#1{\bgroup\markoverwith{\textcolor{purple}{\rule[0.5ex]{2pt}{1pt}}}\ULon{#1}}

\title{Leveraging Large Language Models for Data-to-Text Generation\\for Severely Under-Resourced Languages}
\title{High-quality Data-to-Text Generation for Severely Under-Resourced Languages with Out-of-the-box Large Language Models}

\author{Michela Lorandi \and Anya Belz \\
  ADAPT Research Centre, Dublin City University \\
  \texttt{\{michela.lorandi, anya.belz\}@adaptcentre.ie}}

\begin{document}
\maketitle
\begin{abstract}
The performance of NLP methods for severely under-resourced languages cannot currently hope to match the state of the art in NLP methods for well resourced languages. We explore the extent to which pretrained large language models (LLMs) can bridge this gap, via the example of data-to-text generation for Irish, Welsh, Breton and Maltese. We test LLMs on these under-resourced languages and English, in a range of scenarios. We find that LLMs easily set the state of the art for the under-resourced languages by substantial margins, as measured by both automatic and human evaluations. For all our languages, human evaluation shows on-a-par performance with humans for our best systems, 
but BLEU scores collapse
compared to English, casting doubt on the metric's suitability for evaluating non-task-specific systems. Overall, our results demonstrate the great potential of LLMs to bridge the performance gap for under-resourced languages.


\end{abstract}

\section{Introduction}

\vspace{-.1cm}
Automatically generating text for a given data set (e.g.\ a textual summary) is a much bigger challenge for severely under-resourced languages than  for well resourced languages like English. Creating a rule-based system by hand is one option: slow but faster if language-independent resources can be used \cite{mille-EtAl:2023:WebNLG}. An alternative is task-specific finetuning and collecting training data for it (partly) by hand and/or by collecting/generating silver training data which may be good enough to achieve a desired performance level.

These methods all take varying but considerable amounts of manual work and time. In contrast, using large language models (LLMs) in their `out of the box' state has next to no such overheads. However, at this point their zero-shot ability to generate correct text of sufficient quality 
(e.g.\ in terms of minimum real-world usefulness where first-draft plus post-editing takes less time than from-scratch) for severely under-resourced languages is untested.

Given that by definition LLMs will have seen very little text in under-resourced languages during training, using them in zero-shot mode for text generation in such languages may not seem a promising idea. In this paper, we explore the extent to which it is possible for data-to-text generation, in so doing shedding light on the potential of LLMs to bridge performance gaps between under-resourced languages (the vast majority of the world's languages) and well resourced languages like English.


All code and results are available on GitHub: \url{https://github.com/michelalorandi/D2T-Gen-for-Under-Res-Lang-w-LLMs}.

\section{Related Research}\label{sec:rel-res}

\vspace{-.1cm}
A large number of papers in the past year have reported work on using LLMs, and GPT in particular, in zero or few-shot mode for a wide range of different tasks, including both system development \cite{liu2023deid,long2023large,lu2022learn,wang2023describe,qin2023chatgpt} and evaluation
\cite{chiang-lee-2023-large,wang2022toxicity,chan2023chateval,shen2023large, hada2023large}.

Because the performance of zero-shot LLMs depends on the quality of the prompt, there has been a corresponding flurry of research on prompt engineering, including plan-and-solve prompting \cite{wang2023plan}, tree-of-thought prompting \cite{yao2023tree,long2023large}, and automatic prompt fixing \cite{pearce2023examining}. 


WebNLG 2023 (see below) included a first attempt  \cite{lorandi-belz-2023-data} to perform data-to-text generation for under-resourced languages using out-of-the-box GPT-3.5 plus Google Translate which outperformed other participating systems by considerable margins. We take the same approach but test four LLMs and three MT systems (two closed source and one open source) in a wider range of scenarios, and additionally test our best system on English where the tough state-of-the-art outperforms humans.

\section{Data and Task}\label{sec:webnlg}

\vspace{-.1cm}
WebNLG 2023 is the third iteration of the WebNLG shared task series and focuses on the severely under-resourced European languages Irish, Breton, Welsh and Maltese\footnote{\url{https://synalp.gitlabpages.inria.fr/webnlg-challenge/challenge_2023/}} \cite{cripwell-etal-2023}. The WebNLG 2023 data consists of 1,778 test items 
for each language,  1,399 dev items for Breton, and 1,665 dev items for Welsh, Irish and Maltese. The test sets were manually translated by professional translators from the English originals. Additionally 13,211 training items are provided where texts were automatically translated from English.  

\begin{figure}[h!tb]
\centering
    \includegraphics[width=.4\textwidth]{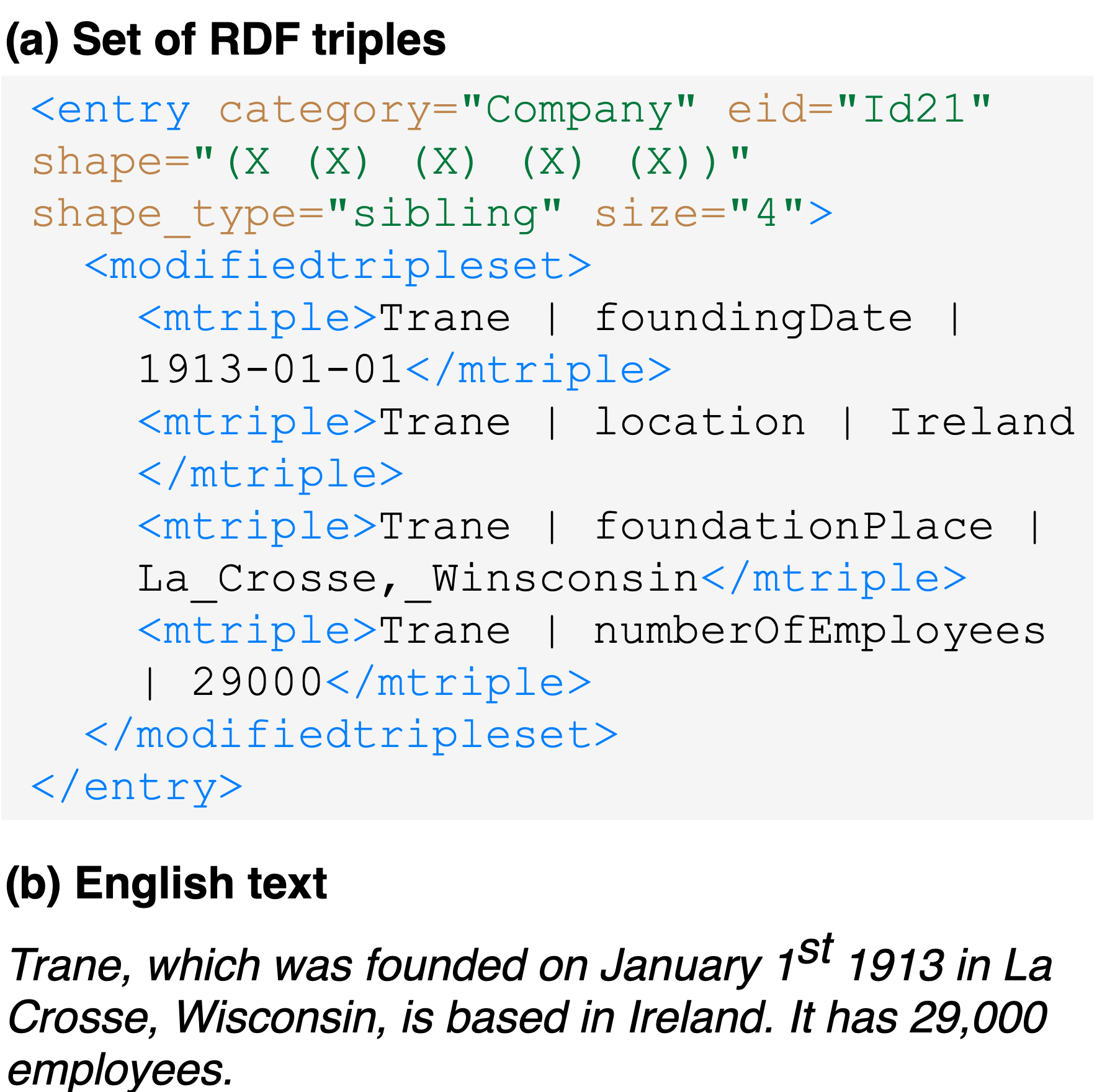}
    \caption{WebNLG input set of triples and output text.}\label{fig:example}
\end{figure}

WebNLG 2023 systems map from RDF triples to a suitable output text, as in the example from the WebNLG'23 website\footnote{\url{https://synalp.gitlabpages.inria.fr/webnlg-challenge/docs}}  in Figure~\ref{fig:example}. The complete shared-task data is available from the same website. 

\section{Models}\label{sec:models}

\vspace{-.1cm}
We test four different pretrained LLMs (paid-for GPT-3.5, and open-source Bloom, LLaMa2-chat, and Falcon-chat), each in two modes: (i) direct generation into the target language, and (ii) generation into English followed by translation into the target language with one of three machine translation (MT) engines (Google Translate, Alibaba Translate, and No Language Left Behind system \cite{costa2022no}). 


\textbf{GPT-3.5} or InstructGPT \cite{ouyang2022training} is GPT-3 plus supervision fine-tuning on instruction  data, reward model training and  Reinforcement Learning with Human Feedback (RLHF) with the reward model.
%
\textbf{BLOOM} \cite{scao2022bloom} is trained on the ROOTS corpus, a collection of 498 HuggingFace datasets.
%
\textbf{LLaMa2-chat} \cite{touvron2023llama} builds on the pretrained LLaMa2 model (trained only on publicly available datasets) fine-tuned in two steps similar to GPT-3.5, 
but instead of using one  reward model for helpfulness and safety, two separately optimised reward models are used.
%
%
\textbf{Falcon-chat} \cite{falcon} builds  on  Falcon-base, which is  trained on the RefinedWeb dataset \cite{penedo2023refinedweb}. Falcon-base is then fine-tuned on chat and instruction datasets with a mix of large-scale conversational datasets.




\section{Experimental Set-up}\label{sec:exp-setup}
\vspace{-.1cm}

In this section we describe the main aspects of the experimental set-up. Hyperparameters and API access are provided in Section~\ref{sec:appendix-hyperparams} in the appendix.

\begin{table*}[h!]
    \centering
    \small
    \setlength\tabcolsep{2pt} 
    \renewcommand{\arraystretch}{1.15}
    \begin{tabular}{|l|l|ccc|ccc|ccc|ccc|}
\hline
      \multirow{2}{*}{\textbf{M}} & \multirow{2}{*}{\textbf{Prompt}} &  \multicolumn{3}{c|}{\textbf{Irish}} &  \multicolumn{3}{c|}{\textbf{Welsh}} &  \multicolumn{3}{c|}{\textbf{Maltese}} &  \multicolumn{3}{c|}{\textbf{Breton}}\\
       & &  \textbf{BLEU}$\uparrow$ & \fontsize{8}{4}\selectfont\textbf{ChrF++}$\uparrow$ & \textbf{TER}$\downarrow$ &  \textbf{BLEU}$\uparrow$ & \fontsize{8}{4}\selectfont \textbf{ChrF++}$\uparrow$ & \textbf{TER}$\downarrow$ &  \textbf{BLEU}$\uparrow$ & \fontsize{8}{4}\selectfont \textbf{ChrF++}$\uparrow$ & \textbf{TER}$\downarrow$ &  \textbf{BLEU}$\uparrow$ & \fontsize{8}{4}\selectfont \textbf{ChrF++}$\uparrow$ & \textbf{TER}$\downarrow$\\
\hline
\multirow{8}{*}{\rotatebox{90}{GPT-3.5 (175B)}} &      ZS MI &  12.9931 &    0.4124 &  0.9298 &  15.8695 &    0.4619 &   0.822  &  13.0311 &     0.445 &  0.8496 &  16.4171 &    0.4303 &  0.7813 \\
            &      FS IC &  15.3477 &    0.4303 &  0.8451 &  18.9512 &    0.4742 &  0.7192 &  15.4315 &    0.4536 &  0.7605 &  \textbf{\textit{18.5925}} &    \textbf{\textit{0.4473}} &  \textbf{\textit{0.7218}} \\
            & ZS MI +GT &  \textbf{\textit{20.5176}} &    \textbf{\textit{0.5146}} &  0.7122 &  24.7126 &    \textbf{\textit{0.5496}} &  0.6659 &  20.3528 &    \textbf{\textit{0.5263}} &    0.67 &        - &         - &       - \\
            & FS IC + GT &  20.4001 &      0.51 &  \textbf{\textit{0.6894}} &   \textbf{\textit{25.115}} &    0.5484 &  \textbf{\textit{0.6435}} &  \textbf{\textit{21.2656}} &    0.5249 &  \textbf{\textit{0.6465}} &        - &         - &       - \\
            & ZS MI + AT &  18.3807 &    0.4984 &  0.7184 &  23.4782 &    0.5408 &  0.6724  &  16.8312 &    0.4902 &    0.72 &  10.5379 &    0.3558 &  0.7954 \\
            & FS IC + AT &  18.3433 &     0.495 &  0.6987 &  23.8908 &    0.5412 &  0.6493  &  17.5723 &    0.4867 &  0.6935 &  10.2411 &    0.3501 &  0.7864 \\
            & ZS+NLLB &  17.5042 &     0.455 &  0.7356 &   19.294 &    0.4761 &  0.6948 &   16.457 &    0.4811 &  0.7262 &      - &      - &     - \\
            & FS+NLLB &  17.1448 &    0.4503 &  0.7136 &   19.106 &    0.4718 &  0.6782 &  17.1262 &     0.479 &  0.7015 &      - &       - &     - \\
\hline
\multirow{8}{*}{\rotatebox{90}{BLOOM (176B)}} &      ZS MI &   2.6099 &    0.2118 &  2.8781 &   1.8576 &    0.2043 &  3.0441 &   2.7287 &    0.2303 &  2.9191 &   1.1293 &     0.161 &  1.8799 \\
            &      FS IC &   4.9828 &    0.2535 &  1.5027 &    6.558 &    0.2696 &  1.1825 &   9.4622 &    0.3075 &  \textit{0.9589} &   5.6066 &    0.2585 &  0.9923 \\
            & ZS MI +GT &   6.6329 &    0.3672 &  2.2041 &   7.4595 &    0.3882 &  2.1584 &   6.3703 &    0.3745 &  2.0717 &        - &         - &       - \\
            & FS IC + GT &  \textit{14.8148} &    \textit{0.4521} &  \textit{0.9073} &  \textit{15.4467} &    \textit{0.4683} &  \textit{0.9699} &  \textit{12.7663} &    \textit{0.4498} &  0.9685 &        - &         - &       - \\
            & ZS MI + AT &   6.2173 &      0.36 &  2.1451 &   7.3117 &    0.3846 &  2.1301 &   5.6348 &    0.3552 &  2.1202 &   4.5007 &    0.2808 &  1.1941 \\
            & FS IC + AT &  12.2466 &    0.4309 &   1.018 &  14.8386 &    0.4621 &  0.9889 &  10.7619 &    0.4229 &  1.0116 &   \textit{8.2509} &    \textit{0.3197} &  \textit{0.8768} \\
            & ZS+NLLB &   4.9851 &    0.2563 &  1.4959 &   5.6246 &    0.2589 &  1.5071 &   4.8973 &    0.2607 &  1.2322 &        - &         - &       - \\
            & FS+NLLB &   7.6891 &    0.2708 &  1.0133 &   8.5701 &    0.2701 &  1.0038 &   6.4824 &    0.2705 &  0.9173 &        - &         - &       - \\
\hline
\multirow{8}{*}{\rotatebox{90}{\fontsize{8}{4}\selectfont LLaMa2-chat (70B)}} &      ZS MI &   6.4367 &    0.2349 &  1.2706 &   6.6383 &    0.2529 &  1.1016 &  10.3055 &    0.3198 &  0.8965 &   4.0113 &    0.2147 &  0.8731 \\
            &      FS IC &  10.4064 &     0.364 &  1.0677 &   8.1874 &    0.3344 &  1.3614 &  12.5935 &    0.3901 &  0.8266 &  \textit{10.2303} &    0.3286 &  0.8095 \\
            & ZS MI +GT &  16.7841 &    0.4872 &  0.8366 &  19.8404 &    0.5212 &  0.8052 &  16.7342 &    0.5028 &  0.7861 &        - &         - &       - \\
            & FS IC + GT &  \textit{19.3366} &    \textit{0.5033} &  \textit{0.7378} &  \textit{23.6408} &    \textit{0.5412} &  \textit{0.6969} &  \textit{19.7145} &    \textit{0.5186} &  \textit{0.6903} &        - &         - &       -\\
            & ZS MI + AT &  16.0344 &    0.4772 &  0.8391 &  19.3043 &    0.5139 &  0.8124 &  13.7873 &     0.471 &  0.8354 &    9.559 &    0.3438 &  0.8448 \\
            & FS IC + AT &  17.9225 &    0.4907 &  0.7458 &  22.5067 &    0.5318 &   0.706 &  15.6232 &    0.4786 &  0.7478 &  10.0142 &    \textit{0.3492} &  \textit{0.8007} \\
            & ZS+NLLB &  15.1903 &    0.4195 &  0.8259 &  16.8335 &    0.4429 &  0.7988 &  14.5649 &    0.4542 &  0.8111 &      - &       - &     - \\
            & FS+NLLB &  16.5713 &     0.442 &  0.7549 &  18.3623 &    0.4632 &  0.7208 &  15.7702 &    0.4648 &  0.7392 &      - &       - &     - \\
\hline
\multirow{8}{*}{\rotatebox{90}{\fontsize{8}{4}\selectfont Falcon-chat (180B)}} &      ZS MI &   6.3239 &    0.2703 &  1.3245 &   6.0496 &    0.2679 &  1.4255 &    6.793 &    0.2765 &  1.3012 &   7.9701 &    0.2638 &   0.923 \\
            &      FS IC &  11.2338 &    0.3657 &  0.9902 &  13.0723 &    0.3611 &  0.8821 &  12.2097 &    0.3656 &  0.8725 &    9.749 &    0.3221 &  0.8079 \\
            & ZS MI +GT &  13.4874 &    0.4584 &  1.1768 &  15.4119 &     0.486 &  1.1724 &  12.9136 &     0.467 &  1.1015 &        - &         - &       - \\
            & FS IC + GT &  \textit{19.6085} &    \textit{0.5034} &  \textit{0.7453} &  \textit{23.1749} &    \textit{0.5387} &  \textit{0.7124} &  \textit{19.5894} &    \textit{0.5158} &  \textit{0.6907} &        - &         - &       -\\
            & ZS MI + AT &  12.5954 &    0.4496 &   1.176 &  14.7283 &    0.4803 &  1.1743 &  10.6168 &    0.4379 &  1.1574 &   8.5235 &    0.3345 &  0.8977 \\
            & FS IC + AT &  17.4847 &    0.4916 &  0.7536 &  22.5094 &    0.5327 &  0.7152 &  15.9008 &    0.4793 &  0.7486 &   \textit{10.285} &    \textit{0.3503} &  \textit{0.8006} \\
            & ZS+NLLB &  12.9335 &    0.4012 &  1.1023 &  13.8666 &    0.4249 &  1.0798 &  11.2754 &    0.4253 &   1.074 &        - &         - &       - \\
            & FS+NLLB &  16.1999 &    0.4385 &  0.7573 &  18.5609 &    0.4631 &  0.7238 &  15.4012 &    0.4623 &    0.74 &        - &         - &       - \\
\hline\hline
\multirow{4}{*}{\rotatebox{90}{\fontsize{8}{4}\selectfont WebNLG23}} & FORGe & 16.66 & 0.44 & 0.75 & - & - & - & - & - & - & - & - & - \\
& IREL     & - & - & - & 20.97 & 0.49 & 0.67 & 16.49 & 0.47 & 0.7 & - & - & - \\
& CUNI-Wue & - & - & - & - & - & - & - & - & - & 10.09 & 0.33 & 0.80 \\
& Baseline & 11.63 & 0.36 & 0.74 & 10.70 & 0.36 & 0.77 & 15.60 & 0.42 & 0.67 & 9.92 & 0.33 & 0.76 \\
\hline
\end{tabular}
    \caption{Automatic evaluation results for \textbf{Irish}, \textbf{Welsh}, \textbf{Maltese} and \textbf{Breton}. Highest score in each column for each language in bold, highest score for each model in italics. Number of parameters in brackets in column~1. ZS MI=Zero-Shot Minimal Instruction, FS IC=Few-Shot In Context, GT=Google Translate, AT=Alibaba Translate, NLLB=No Language Left Behind system.}
    \label{tab:ga_cy_res}
\end{table*}

\subsection{Experimental grid}\label{ssec:grid}
\vspace{-.05cm}

We tested all combinations of our four LLMs, two translation engines, two prompts, and five languages, i.e.\ the basic experimental grid looks as follows: 
    \{GPT-3.5, Bloom, Llama2, Falcon\} $\times$ \{Google Translate, Alibaba Translate, NLLB system\} $\times$ \{zero-shot minimal instruction, few-shot in context\} $\times$ \{Irish, Breton, Maltese, Welsh, English\}.

\subsection{Prompt engineering}
\vspace{-.05cm}

We use the prompts previously identified  \cite{lorandi-belz-2023-data} as the most suitable for data-to-text generation following prompt testing of {zero-shot minimal instruction}, {few-shot in-context learning}, and {chain-of-thought} (CoT) \cite{wei2022chain} on GPT-3.5 and GPT-4, on a 
different random sample of 20 data/text pairs in each phase.


For the work reported here, we conducted a preliminary testing phase with BLOOM, LLaMa2, and Falcon to verify if further postprocessing is needed. As a result, we remove all Python code, occurrences of """, and output start markers (e.g.\ "Falcon:") from the output of all three. 

\subsection{Evaluation}
\vspace{-.05cm}

We carried out automatic evaluations with BLEU \cite{papineni2002bleu}, ChrF++ \cite{popovic-2017-chrf} and TER \cite{snover-etal-2006-study} for all systems (each cell in the experimental grid from Section~\ref{ssec:grid}); the resulting scores are shown in Table~\ref{tab:ga_cy_res}. Furthermore, we computed COMET \cite{rei2020comet} for all systems, and BERTScore \cite{zhang2019bertscore} for all Irish, Welsh and Breton systems (see Appendix \ref{sec:appendix-res}). 

We report a new human evaluation of four of the English systems using exactly the same method as in WebNLG 2023 \cite{cripwell-etal-2023}. In terms of the experimental grid above, the four systems in the human evaluation were \{GPT-3.5\} $\times$ \{\} $\times$ \{zero-shot minimal instruction, few-shot in context\} $\times$ \{English\}. We evaluated these alongside the best English system from WebNLG 2020, and the human-authored test-set outputs.

We also  include relevant results from the WebNLG 2020 and 2023 human evaluations, from the latter for \{GPT-3.5\} $\times$ \{Google Translate\} $\times$ \{few-shot in context\} $\times$ \{Irish, Maltese, Welsh\}, and the second best WebNLG 2023 system.

\section{Results}\label{sec:results}
\vspace{-.1cm}

This section reports the main human and metric evaluation results. Details of cost in Section~\ref{sec:appendix-cost}.

\subsection{Metrics}
\vspace{-.05cm}

Metric results (BLEU, ChrF++ and TER) for all systems in our grid from Section~\ref{ssec:grid} are shown in Table~\ref{tab:ga_cy_res} for Irish/Welsh/ Maltese/Breton, and in Table~\ref{tab:en_res} for English. Tables~\ref{tab:bert-res},~\ref{tab:res-en-bert} and~\ref{tab:res-comet} present BERTScore and COMET metric results for Irish/Welsh/Breton, English, and Irish/Welsh/Maltese/Breton/English, respectively. 

High-level results across all languages are that GPT-3.5+GoogleTrans always has a higher metric score than all other model/translation engine combinations, except for English where it has the highest score for ChrF++, but is outperformed by the top-ranking WebNLG 2020 system for BLEU and TER. 

Generation into English plus Google Translate has better scores than direct generation into the under-resourced language by substantial margins in all cases. Alibaba has slightly better scores than direct generation in  all cases except Breton, while NLLB has slightly better scores than direct generation, but worse than Alibaba, in the majority of cases. 

For all models except GPT, the few-shot version of a system is always better than the zero-shot. For GPT the few-shot and zero-shot results are much closer, and in a few cases, zero-shot is  slightly better than few shot, e.g.\ for Maltese using translation.

For the under-resourced languages, the overall best metric scores are obtained for Welsh, by good margins, followed by Maltese, Irish, and Breton, where we cannot use Google Translate, and where in fact generation into English plus Alibaba is a lot worse than direct generation in case of GPT-3.5. This is in contrast to the other languages where Alibaba always achieves small improvements.

Considering COMET (Table~\ref{tab:res-comet}), we get similar results for GPT-3.5 and Falcon-chat when using a MT system and Few-Shot In-Context prompt in all under-resourced languages. 

\begin{table}[h!]
    \centering
    \small
    \setlength\tabcolsep{4.5pt} 
    \renewcommand{\arraystretch}{1.15}
    \begin{tabular}{|ll|ccc|}
        \hline
        \textbf{Model} & \textbf{Prompt} & \textbf{BLEU} $\uparrow$ & \textbf{ChrF++} $\uparrow$ & \textbf{TER} $\downarrow$ \\
        \hline
GPT-3.5 &  ZS MI &  49.6603 &  0.6895 &  0.4498 \\
(175B) &  FS IC &  \textit{52.7366} &  \textbf{\textit{0.6906}} &    \textit{0.42} \\
         \hline
BLOOM &  ZS MI &  13.4535 &  0.4572 &   0.705 \\
(176B) &  FS IC &  \textit{32.1397} &  \textit{0.5816} &  \textit{0.5876} \\
         \hline
LLaMa2-chat &  ZS MI &  40.4711 &  0.6421 &  0.5746 \\
(70B) &  FS IC &  \textit{46.8566} &  \textit{0.6705} &  \textit{0.4853} \\
         \hline
Falcon-chat &  ZS MI &  31.3463 &  0.5922 &  0.6545 \\
(180B) &  FS IC &  \textit{46.3762} &   \textit{0.668} &  \textit{0.4891} \\
         \hline \hline
         \multicolumn{2}{|l|}{WebNLG 2020:} &  &  &  \\
         \multicolumn{2}{|l|}{Baseline FORGE2020} & 40.6 & 62.1 & 51.7 \\
         \multicolumn{2}{|l|}{Amazon AI (Shanghai)} & \textbf{54.0} & {69.0} & \textbf{40.6} \\
         \multicolumn{2}{|l|}{OSU Neural NLG} & 53.5 & 68.8 & 41.6 \\
        \hline
    \end{tabular}
    \caption{Automatic evaluation results for \textbf{English}. Best score per column in bold, best score per model in italics. Number of model parameters in brackets. ZS MI=Zero-Shot Minimal Instruction, FS IC=Few-Shot In Context.}
    \label{tab:en_res}
\end{table}

An interesting aspect of the metric results is that while best BLEU scores are far higher for English than for any other language (e.g.\ more than twice as high for the best results), this pattern is not replicated in the ChrF++, TER, BERTScore and COMET scores. See Section~\ref{sec:disc-concl} for discussion.

\newcommand{\cuni}{{CUNI-Wue}}%
\newcommand{\mille}{{DCU/TCD-FORGe}} 
\newcommand{\interno}{{Interno}}
\newcommand{\irel}{{IREL}}
\newcommand{\lorandi}{{DCU-NLG-PBN}}
\newcommand{\wnlg}{{WebNLG}}


\subsection{Human evaluation of English systems}
\vspace{-.05cm}


Outside of WebNLG 2023, there is no state of the art for data-to-text generation in our four under-resourced languages that we can compare against. However, we can compare our methods against the best performing systems in English from WebNLG 2020, and we did this using the  same human evaluation approach that was used in WebNLG 2023.

Table~\ref{tab:english-hum-eval} shows the results from this evaluation of Fluency, Absence of Additions, and Absence of Omissions 
which show that  few-shot GPT3.5  has the highest mean score for Fluency, Omissions and Repetition, with zero-shot having the highest mean in Additions. However, there are significant performance differences only for Omissions, reflecting a similar relatively lower score for Omissions in the WebNLG20 evaluations (see next section).

\begin{table}[h!]
    \centering
    \small
    \setlength\tabcolsep{5.6pt} 
    \renewcommand{\arraystretch}{1.15}
    \begin{tabular}{|l|cc|cc|ccc|}
\hline
          \textbf{System} &  \multicolumn{2}{c|}{\textbf{Fluency}} &  \multicolumn{2}{c|}{\textbf{Addition}} &  \multicolumn{3}{c|}{\textbf{Omission}} \\
\hline
GPT-3.5 FS MI &      \textbf{4.50} &    A &      0.88 &    A &      \textbf{0.93} & A &      \\
Amazon AI &      4.33 &    A &      0.90 &    A &      0.82 &  & B  \\
GPT-3.5 ZS IC &      4.33 &    A &      \textbf{0.91} &    A &      \textbf{0.93} & A & \\
Human ref &      4.28 &    A &      0.83 &    A &      0.92 &    A &    B  \\
\hline
\end{tabular}
    \caption{Human evaluation results for \textbf{English} for  human-authored references, GPT-3.5 zero-shot, GPT-3.5 few-shot), and best WebNLG20 system. Means and homogeneous subsets from Tukey HSD (alpha = .05).}
    \label{tab:english-hum-eval}
\end{table}

\subsection{WebNLG human evaluations}
\vspace{-.05cm}


Table~\ref{tab:webnlg23_hum_eval}  shows mean \textbf{WebNLG 2023} human scores for \textbf{Welsh}, \textbf{Maltese} and \textbf{Irish}, per system for Fluency, Addition and Omission, for the human reference texts, the GPT-3.5+Google Translate+few-shot system (DCU-NLG-PBN) and the next best system. 
\vspace{.01cm}

\begin{table}[h!]
    \setlength\tabcolsep{1.9pt} 
    \renewcommand{\arraystretch}{1.15}
    \begin{small}
    \begin{tabular}{|l|l|cccc|cccc|ccc|}
        \hline
        \textbf{L} & \textbf{System} & \multicolumn{4}{c|}{\textbf{Fluency}} & \multicolumn{4}{c|}{\textbf{Addition}} & \multicolumn{3}{c|}{\textbf{Omission}} \\
        \hline
        
        \multirow{3}{*}{\rotatebox[]{90}{Welsh}} & Human ref & \textbf{3.28} & A & & & \textbf{0.9} & A & & & \textbf{0.84} & A & \\
         & \fontsize{8}{4}\selectfont {DCU-NLG-PBN}& 3.25 & A &  &  & 0.86 & A &  &  & 0.77 & A &  \\
         & \irel & 2.67 &  & B &  & 0.6 &  & B &  & 0.47 &  & B \\
         
         \hline

         \multirow{3}{*}{\rotatebox[]{90}{Maltese}} & Human ref & \textbf{4.27} & A & & & 0.89 & A & & & 0.85 & A & \\
         & \fontsize{8}{4}\selectfont {DCU-NLG-PBN} & 4.06 & A & B &  & \textbf{0.91} & A &  &  & \textbf{0.86} & A & \\
         & \irel & 3.74 &  & B &  & 0.69 &  &  B&  & 0.56 &  & B \\

         \hline
         
        \multirow{3}{*}{\rotatebox[]{90}{Irish}} & Human ref & \textbf{4.07} & A & & & 0.81 & A & & & 0.82 & A & \\
         & \fontsize{8}{4}\selectfont DCU-NLG-PBN& 3.83 & A & B &  & 0.83 & A &  &  & \textbf{0.85} & A & \\
         & \fontsize{7.5}{4}\selectfont DCU/TCD-FORGe & 3.35 &  &  & C & \textbf{0.84} & A &  &  & 0.81 & A & \\
         \hline 
    \end{tabular}
    \end{small}
    \caption{Mean \textbf{WebNLG 2023} human scores for \textbf{Welsh}, \textbf{Maltese} and \textbf{Irish}, per system for Fluency, Addition and Omission.}
    \label{tab:webnlg23_hum_eval}
\end{table}
\vspace{.01cm}

Here too, the differences between the scores for the human references and the DCU-NLG-PGN system (few-shot GPT + GT) are not statistically significant for any of the nine sets of scores; the human references come top 5 times, DCU-NLG-PGN 3 times, and DCU/TCD-FORGe once. The human references and the DCU-NLG-PBN system are significantly better than the runner up system for Maltese and Welsh on all evaluation criteria. Taken together, we can consider that on-par-with-human performance for the GPT+MT systems.

In Table~\ref{tab:webnlg2020_hum_eval}, we show results for the \textbf{English} human evaluation from \textbf{WebNLG 2020} for reference (evaluation criteria translated to match our terminology).
\vspace{.01cm}

\begin{table}[h!]
    \centering
    \setlength\tabcolsep{2.8pt} 
    \renewcommand{\arraystretch}{1.15}
    \begin{small}
    \begin{tabular}{|l|l|cc|cc|cc|}
        \hline
        \textbf{L} & \textbf{System} & \multicolumn{2}{c|}{\textbf{Fluency}} & \multicolumn{2}{c|}{\textbf{Addition}} & \multicolumn{2}{c|}{\textbf{Omission}} \\
        \hline
        
        \multirow{3}{*}{\rotatebox[]{90}{English}} & Amazon AI & \textbf{90.286} & A &	\textbf{95.196} & A &	94.393 & A \\
         & \fontsize{8}{4}\selectfont {OSU Neural NLG}& 90.066 & A &	94.615  & A &	95.123 & A \\
         & Human ref & 89.846 & A & 94.392 & A & \textbf{95.442} & A \\
         \hline
    \end{tabular}
    \end{small}
    \caption{Human evaluation results of \textbf{English} from \textbf{WebNLG 2020}.}
    \label{tab:webnlg2020_hum_eval}
\end{table}
\vspace{.01cm}


\noindent The two systems have slightly higher scores than the human references except for Omissions. Recall that Table~\ref{tab:english-hum-eval} indicates that GPT3.5+MT outperforms the Amazon AI system and the human references. Taken together the two human evaluations indicate overall better performance for GPT3.5+MT.

\section{Discussion and Conclusion}\label{sec:disc-concl}
\vspace{-.1cm}

One striking aspect of the metric results for the under-resourced languages is that BLEU scores are far lower across the board than for English. At the same time, human evaluations show on-a-par-with-human performance for both the under-resourced languages and English. This shows a significant performance failure for BLEU that is not reflected in ChrF++, TER, BERTScore or COMET. 

This BLEU failure may be due to two aspects: for one, BLEU is a word n-gram overlap metric, while ChrF++ and TER are character F-Score and character edit distance based, respectively. BERTScore computes cosine similarity for each token in candidate and reference sentences using the pre-trained contextual embeddings from BERT, and COMET uses a pretrained multilingual model trained to mimic human judgement. Two, the GPT training data is likely to have contained the English WebNLG data in its entirety (albeit not as input/output pairs), but not any of the under-resourced language outputs. It seems that under these circumstances, where system outputs and reference texts have not been sampled from the same narrow distribution, BLEU simply does not work.

The systems that we introduce and test here are 
generic, non-task-specifically trained systems. All of the systems we compare them against are task-specifically supervision-trained systems, and in one case \cite{mille-EtAl:2023:WebNLG}, hand-crafted to perform a single specific task. It is yet another piece of evidence showcasing the astonishing out-of-the-box abilities of the latest generation of LLMs. Similarly to previous evidence, we see that absence of instruction tuning (BLOOM) and smaller size (LLaMa2) are associated with poorer performance. It is also unclear how such systems can be utilised in real-world application scenarios. However, we show the incredible ability of LLMs to generate texts on-a-par performance with humans for
our best systems in all languages tested.





\section*{Limitations}
In this work, we focused on the usage of LLMs together with MT engines. Not all the models used are open-sourced and to access them we need to use paid APIs. This not only implies a financial cost that could be prohibited, but also implies problems in terms of reproducibility as we're not entirely sure of what the model is behind the APIs.

Furthermore, considering the open-sourced LLMs, we need a large number of GPUs to be able to execute such models, especially BLOOM (176B) and Falcon (180B). In the case of Falcon, we would need at least 400GB of memory to run the model in inference.

Lastly, we explored only two simple types of prompts designed based on GPT-3.5 and it could be beneficial to explore more advanced types of prompts also taking into account differences between models.

\section*{Ethics Statement}
We focused on under-resourced languages setting a base for further research and the development of real-world applications that people who speak such languages could use. 
On the other hand, when using LLMs there is a general risk that they could produce offensive or incorrect content that may harm people using such systems. 
Since our approach only takes into account the given input without any factual checking, we cannot guarantee that there is no generation of factually incorrect texts.

Furthermore, it's currently unclear what has been included in the training data of some LLMs, meaning that there may be evidence of bias in generated texts, which in turn carries a risk of possibly causing harm to the end user.

\section*{Acknowledgements}
The cost of accessing the GPT API was covered by financial support from the DCU-NLG Research Group at DCU. Michela Lorandi's work was conducted with the financial support of the Science Foundation Ireland Centre for Research Training in Digitally-Enhanced Reality (d-real) under Grant No. 18/CRT/6224. Both authors benefit from being members of the ADAPT SFI Research Centre at Dublin City University, funded by the Science Foundation Ireland under Grant Agreement No. 13/RC/2106\_P2. For the purpose of Open Access, the author has applied a CC BY public copyright licence to any Author Accepted Manuscript version arising from this submission.

\bibliography{anthology,custom}

\begin{thebibliography}{28}
\expandafter\ifx\csname natexlab\endcsname\relax\def\natexlab#1{#1}\fi

\bibitem[{Almazrouei et~al.(2023)Almazrouei, Alobeidli, Alshamsi, Cappelli, Cojocaru, Alhammadi, Daniele, Heslow, Launay, Malartic, Noune, Pannier, and Penedo}]{falcon}
Ebtesam Almazrouei, Hamza Alobeidli, Abdulaziz Alshamsi, Alessandro Cappelli, Ruxandra Cojocaru, Maitha Alhammadi, Mazzotta Daniele, Daniel Heslow, Julien Launay, Quentin Malartic, Badreddine Noune, Baptiste Pannier, and Guilherme Penedo. 2023.
\newblock The falcon series of language models: Towards open frontier models.

\bibitem[{Chan et~al.(2023)Chan, Chen, Su, Yu, Xue, Zhang, Fu, and Liu}]{chan2023chateval}
Chi-Min Chan, Weize Chen, Yusheng Su, Jianxuan Yu, Wei Xue, Shanghang Zhang, Jie Fu, and Zhiyuan Liu. 2023.
\newblock Chateval: Towards better llm-based evaluators through multi-agent debate.
\newblock \emph{arXiv preprint arXiv:2308.07201}.

\bibitem[{Chiang and Lee(2023)}]{chiang-lee-2023-large}
Cheng-Han Chiang and Hung-yi Lee. 2023.
\newblock \href {https://doi.org/10.18653/v1/2023.acl-long.870} {Can large language models be an alternative to human evaluations?}
\newblock In \emph{Proceedings of the 61st Annual Meeting of the Association for Computational Linguistics (Volume 1: Long Papers)}, pages 15607--15631, Toronto, Canada. Association for Computational Linguistics.

\bibitem[{Costa-juss{\`a} et~al.(2022)Costa-juss{\`a}, Cross, {\c{C}}elebi, Elbayad, Heafield, Heffernan, Kalbassi, Lam, Licht, Maillard et~al.}]{costa2022no}
Marta~R Costa-juss{\`a}, James Cross, Onur {\c{C}}elebi, Maha Elbayad, Kenneth Heafield, Kevin Heffernan, Elahe Kalbassi, Janice Lam, Daniel Licht, Jean Maillard, et~al. 2022.
\newblock No language left behind: Scaling human-centered machine translation.
\newblock \emph{arXiv preprint arXiv:2207.04672}.

\bibitem[{Cripwell et~al.(2023)Cripwell, , Belz, Borg, Gardent, Gatt, Judge, Lorandi, Nikiforoskaya, Soto-Martinez, and Thomson}]{cripwell-etal-2023}
Liam Cripwell, , Anya Belz, Claudia Borg, Claire Gardent, Albert Gatt, John Judge, Michela Lorandi, Anna Nikiforoskaya, William Soto-Martinez, and Craig Thomson. 2023.
\newblock The 2023 webnlg shared task on low resource languages overview and evaluation results (webnlg 2023).
\newblock In \emph{Proceedings of the Workshop on Multimodal, Multilingual Natural Language Generation and Multilingual WebNLG Challenge}, Prague, Czech Republic.

\bibitem[{Hada et~al.(2023)Hada, Gumma, de~Wynter, Diddee, Ahmed, Choudhury, Bali, and Sitaram}]{hada2023large}
Rishav Hada, Varun Gumma, Adrian de~Wynter, Harshita Diddee, Mohamed Ahmed, Monojit Choudhury, Kalika Bali, and Sunayana Sitaram. 2023.
\newblock Are large language model-based evaluators the solution to scaling up multilingual evaluation?
\newblock \emph{arXiv preprint arXiv:2309.07462}.

\bibitem[{Liu et~al.(2023)Liu, Yu, Zhang, Wu, Cao, Dai, Zhao, Liu, Shen, Li et~al.}]{liu2023deid}
Zhengliang Liu, Xiaowei Yu, Lu~Zhang, Zihao Wu, Chao Cao, Haixing Dai, Lin Zhao, Wei Liu, Dinggang Shen, Quanzheng Li, et~al. 2023.
\newblock Deid-gpt: Zero-shot medical text de-identification by gpt-4.
\newblock \emph{arXiv preprint arXiv:2303.11032}.

\bibitem[{Long(2023)}]{long2023large}
Jieyi Long. 2023.
\newblock Large language model guided tree-of-thought.
\newblock \emph{arXiv preprint arXiv:2305.08291}.

\bibitem[{Lorandi and Belz(2023)}]{lorandi-belz-2023-data}
Michela Lorandi and Anya Belz. 2023.
\newblock \href {https://aclanthology.org/2023.mmnlg-1.9} {Data-to-text generation for severely under-resourced languages with {GPT}-3.5: A bit of help needed from {G}oogle {T}ranslate ({W}eb{NLG} 2023)}.
\newblock In \emph{Proceedings of the Workshop on Multimodal, Multilingual Natural Language Generation and Multilingual WebNLG Challenge (MM-NLG 2023)}, pages 80--86, Prague, Czech Republic. Association for Computational Linguistics.

\bibitem[{Lu et~al.(2022)Lu, Mishra, Xia, Qiu, Chang, Zhu, Tafjord, Clark, and Kalyan}]{lu2022learn}
Pan Lu, Swaroop Mishra, Tanglin Xia, Liang Qiu, Kai-Wei Chang, Song-Chun Zhu, Oyvind Tafjord, Peter Clark, and Ashwin Kalyan. 2022.
\newblock Learn to explain: Multimodal reasoning via thought chains for science question answering.
\newblock \emph{Advances in Neural Information Processing Systems}, 35:2507--2521.

\bibitem[{Mille et~al.(2023)Mille, U'i~Dhonnchadha, Dasiopoulou, Cassidy, Davis, and Belz}]{mille-EtAl:2023:WebNLG}
Simon Mille, Elaine U'i~Dhonnchadha, Stamatia Dasiopoulou, Lauren Cassidy, Brian Davis, and Anya Belz. 2023.
\newblock {DCU}/{TCD}-{FORG}e at {W}eb{NLG}’23: {I}rish rules!
\newblock In \emph{Proceedings of the Workshop on Multimodal, Multilingual Natural Language Generation and Multilingual WebNLG Challenge}, Prague, Czech Republic.

\bibitem[{Ouyang et~al.(2022)Ouyang, Wu, Jiang, Almeida, Wainwright, Mishkin, Zhang, Agarwal, Slama, Ray et~al.}]{ouyang2022training}
Long Ouyang, Jeffrey Wu, Xu~Jiang, Diogo Almeida, Carroll Wainwright, Pamela Mishkin, Chong Zhang, Sandhini Agarwal, Katarina Slama, Alex Ray, et~al. 2022.
\newblock Training language models to follow instructions with human feedback.
\newblock \emph{Advances in Neural Information Processing Systems}, 35:27730--27744.

\bibitem[{Papineni et~al.(2002)Papineni, Roukos, Ward, and Zhu}]{papineni2002bleu}
Kishore Papineni, Salim Roukos, Todd Ward, and Wei-Jing Zhu. 2002.
\newblock Bleu: a method for automatic evaluation of machine translation.
\newblock In \emph{Proceedings of the 40th annual meeting of the Association for Computational Linguistics}, pages 311--318.

\bibitem[{Pearce et~al.(2023)Pearce, Tan, Ahmad, Karri, and Dolan-Gavitt}]{pearce2023examining}
Hammond Pearce, Benjamin Tan, Baleegh Ahmad, Ramesh Karri, and Brendan Dolan-Gavitt. 2023.
\newblock Examining zero-shot vulnerability repair with large language models.
\newblock In \emph{2023 IEEE Symposium on Security and Privacy (SP)}, pages 2339--2356. IEEE.

\bibitem[{Penedo et~al.(2023)Penedo, Malartic, Hesslow, Cojocaru, Cappelli, Alobeidli, Pannier, Almazrouei, and Launay}]{penedo2023refinedweb}
Guilherme Penedo, Quentin Malartic, Daniel Hesslow, Ruxandra Cojocaru, Alessandro Cappelli, Hamza Alobeidli, Baptiste Pannier, Ebtesam Almazrouei, and Julien Launay. 2023.
\newblock The refinedweb dataset for falcon llm: outperforming curated corpora with web data, and web data only.
\newblock \emph{arXiv preprint arXiv:2306.01116}.

\bibitem[{Popovi{\'c}(2017)}]{popovic-2017-chrf}
Maja Popovi{\'c}. 2017.
\newblock \href {https://doi.org/10.18653/v1/W17-4770} {chr{F}++: words helping character n-grams}.
\newblock In \emph{Proceedings of the Second Conference on Machine Translation}, pages 612--618, Copenhagen, Denmark. Association for Computational Linguistics.

\bibitem[{Qin et~al.(2023)Qin, Zhang, Zhang, Chen, Yasunaga, and Yang}]{qin2023chatgpt}
Chengwei Qin, Aston Zhang, Zhuosheng Zhang, Jiaao Chen, Michihiro Yasunaga, and Diyi Yang. 2023.
\newblock Is chatgpt a general-purpose natural language processing task solver?
\newblock \emph{arXiv preprint arXiv:2302.06476}.

\bibitem[{Rei et~al.(2020)Rei, Stewart, Farinha, and Lavie}]{rei2020comet}
Ricardo Rei, Craig Stewart, Ana~C Farinha, and Alon Lavie. 2020.
\newblock Comet: A neural framework for mt evaluation.
\newblock In \emph{Proceedings of the 2020 Conference on Empirical Methods in Natural Language Processing (EMNLP)}, pages 2685--2702.

\bibitem[{Scao et~al.(2022)Scao, Fan, Akiki, Pavlick, Ili{\'c}, Hesslow, Castagn{\'e}, Luccioni, Yvon, Gall{\'e} et~al.}]{scao2022bloom}
Teven~Le Scao, Angela Fan, Christopher Akiki, Ellie Pavlick, Suzana Ili{\'c}, Daniel Hesslow, Roman Castagn{\'e}, Alexandra~Sasha Luccioni, Fran{\c{c}}ois Yvon, Matthias Gall{\'e}, et~al. 2022.
\newblock Bloom: A 176b-parameter open-access multilingual language model.
\newblock \emph{arXiv preprint arXiv:2211.05100}.

\bibitem[{Shen et~al.(2023)Shen, Cheng, You, and Bing}]{shen2023large}
Chenhui Shen, Liying Cheng, Yang You, and Lidong Bing. 2023.
\newblock Are large language models good evaluators for abstractive summarization?
\newblock \emph{arXiv preprint arXiv:2305.13091}.

\bibitem[{Snover et~al.(2006)Snover, Dorr, Schwartz, Micciulla, and Makhoul}]{snover-etal-2006-study}
Matthew Snover, Bonnie Dorr, Rich Schwartz, Linnea Micciulla, and John Makhoul. 2006.
\newblock \href {https://aclanthology.org/2006.amta-papers.25} {A study of translation edit rate with targeted human annotation}.
\newblock In \emph{Proceedings of the 7th Conference of the Association for Machine Translation in the Americas: Technical Papers}, pages 223--231, Cambridge, Massachusetts, USA. Association for Machine Translation in the Americas.

\bibitem[{Touvron et~al.(2023)Touvron, Martin, Stone, Albert, Almahairi, Babaei, Bashlykov, Batra, Bhargava, Bhosale et~al.}]{touvron2023llama}
Hugo Touvron, Louis Martin, Kevin Stone, Peter Albert, Amjad Almahairi, Yasmine Babaei, Nikolay Bashlykov, Soumya Batra, Prajjwal Bhargava, Shruti Bhosale, et~al. 2023.
\newblock Llama 2: Open foundation and fine-tuned chat models.
\newblock \emph{arXiv preprint arXiv:2307.09288}.

\bibitem[{Wang et~al.(2023{\natexlab{a}})Wang, Xu, Lan, Hu, Lan, Lee, and Lim}]{wang2023plan}
Lei Wang, Wanyu Xu, Yihuai Lan, Zhiqiang Hu, Yunshi Lan, Roy Ka-Wei Lee, and Ee-Peng Lim. 2023{\natexlab{a}}.
\newblock Plan-and-solve prompting: Improving zero-shot chain-of-thought reasoning by large language models.
\newblock \emph{arXiv preprint arXiv:2305.04091}.

\bibitem[{Wang and Chang(2022)}]{wang2022toxicity}
Yau-Shian Wang and Yingshan Chang. 2022.
\newblock Toxicity detection with generative prompt-based inference.
\newblock \emph{arXiv preprint arXiv:2205.12390}.

\bibitem[{Wang et~al.(2023{\natexlab{b}})Wang, Cai, Liu, Ma, and Liang}]{wang2023describe}
Zihao Wang, Shaofei Cai, Anji Liu, Xiaojian Ma, and Yitao Liang. 2023{\natexlab{b}}.
\newblock Describe, explain, plan and select: Interactive planning with large language models enables open-world multi-task agents.
\newblock \emph{arXiv preprint arXiv:2302.01560}.

\bibitem[{Wei et~al.(2022)Wei, Wang, Schuurmans, Bosma, Chi, Le, and Zhou}]{wei2022chain}
Jason Wei, Xuezhi Wang, Dale Schuurmans, Maarten Bosma, Ed~Chi, Quoc Le, and Denny Zhou. 2022.
\newblock Chain of thought prompting elicits reasoning in large language models.
\newblock \emph{arXiv preprint arXiv:2201.11903}.

\bibitem[{Yao et~al.(2023)Yao, Yu, Zhao, Shafran, Griffiths, Cao, and Narasimhan}]{yao2023tree}
Shunyu Yao, Dian Yu, Jeffrey Zhao, Izhak Shafran, Thomas~L Griffiths, Yuan Cao, and Karthik Narasimhan. 2023.
\newblock Tree of thoughts: Deliberate problem solving with large language models.
\newblock \emph{arXiv preprint arXiv:2305.10601}.

\bibitem[{Zhang et~al.(2019)Zhang, Kishore, Wu, Weinberger, and Artzi}]{zhang2019bertscore}
Tianyi Zhang, Varsha Kishore, Felix Wu, Kilian~Q Weinberger, and Yoav Artzi. 2019.
\newblock Bertscore: Evaluating text generation with bert.
\newblock \emph{arXiv preprint arXiv:1904.09675}.

\end{thebibliography}
\bibliographystyle{acl_natbib}

\appendix

\section{Appendix}

\subsection{Hyperparameters and APIs}
\label{sec:appendix-hyperparams}

We executed all the experiments either via API or on our own GPUs. We used the paid-for OpenAI API to access text-davinci-003 \footnote{\url{https://platform.openai.com/docs/models/gpt-3-5}} (GPT-3.5), while we used the free inference API of HuggingFace to access BLOOM 176B \footnote{\url{https://huggingface.co/bigscience/bloom}} and falcon-180B-chat \footnote{\url{https://huggingface.co/tiiuae/falcon-180B-chat}}. On the other hand, we downloaded and executed Llama-2-70b-chat-hf\footnote{\url{https://huggingface.co/meta-llama/Llama-2-70b-chat-hf}} on a Nvidia A100 GPU with 80GB RAM. 

To use the three explored Machine Translation engines, we used the pay-as-you-go APIs of Google Cloud \footnote{\url{https://cloud.google.com/translate}} and Alibaba Cloud \footnote{\url{https://www.alibabacloud.com/product/machine-translation}}, and we downloaded and executed NLLB \cite{costa2022no} on a Nvidia A100 GPU with 80GB RAM.

For all used models, we set \textit{maximum length} to 500 with Zero-Shot Minimal Instruction and 1000 with Few-Shot In Context. All generated texts are post-processed as described above. 

\paragraph{GPT-3.5} In all experiments involving GPT-3.5, we set text-davinci-003 parameters to \textit{temperature}=0, \textit{top p}=1 (default), \textit{frequency penalty}=0 and \textit{presence penalty}=0 (default), \textit{best of}=1 (default) to get only 1 completion for each prompt.

\paragraph{BLOOM} We used bigscience/bloom model with HuggingFace's Inference Client API setting the parameters to \textit{temperature}=0.7, \textit{top p}=0.9, \textit{frequency penalty}=0 and \textit{presence penalty}=0.

\paragraph{LLaMa2-chat} We used meta-llama/Llama-2-70b-chat-hf model on HuggingFace setting the parameters to \textit{temperature}=1 (default), \textit{top p}=1 (default), \textit{repetition penalty}=1 (default) and \textit{diversity penalty}=0 (default), \textit{num return sequences}=1.

\paragraph{Falcon-chat} We used tiiuae/falcon-180b-chat model with HuggingFace's Inference Client API setting the parameters to \textit{temperature}=0.7, \textit{top p}=0.9, \textit{frequency penalty}=0 and \textit{presence penalty}=0.

\paragraph{NLLB} We used facebook/nllb-200-1.3B model on HuggingFace setting the languages to \textit{mlt\_Latn}, \textit{cym\_Latn}, and \textit{gle\_Latn}, respectively for Maltese, Welsh, and Irish.

\paragraph{COMET} We used the Unbabel/wmt22-comet-da model on HuggingFace.

\subsection{Computational and financial cost}\label{sec:appendix-cost}
To execute our experiments, we relied on the use of paid APIs and GPU usage. 

Considering paid APIs, GPT-3.5 model cost US\$91.82 in API, while the usage of Google Translate and Alibaba cost respectively €135.15 and US\$377.97.

Regarding computational time and cost, we executed all LLama2 chat experiments on a Nvidia A100 GPU, which took, on average, around 21 hours to execute a single experiment using Zero-Shot Minimal Instruction (ZS MI) prompt and around 2 days and 18 hours to execute a single experiment using Few-Shot In Context (FS IC) prompt. On the other hand, we accessed all the other models through API calls. On average, using HuggingFace inference API BLOOM176B took around 17 hours for ZS MI prompt and around 2 days for FS CI prompt, while Falcon 180B took around 11 hours for ZS MI prompt and around 20 hours for FS CI prompt. Lastly, using GPT-3.5 with OpenAI APIs, it took around 1 hour both for ZS MI and FS CI prompts.

\section{Additional results}
\label{sec:appendix-res}

In this Section, we provide additional automatic evaluation results using BERTScore and COMET.

Tables \ref{tab:bert-res} and \ref{tab:res-en-bert} present BERTScore results for all systems in  Irish/Welsh/Breton and English, respectively. Maltese is not included as it is not available in BERTScore.

Tables \ref{tab:res-comet} present COMET results for all systems in our grid from Section \ref{ssec:grid}, for Irish/Welsh/Maltese/Breton/English.

\begin{table*}[h!tb]
    \centering
    \small
    \setlength\tabcolsep{2pt} 
    \renewcommand{\arraystretch}{1.15}
    \begin{tabular}{|l|l|ccc|ccc|ccc|ccc|}
\hline
      \multirow{2}{*}{\textbf{M}} & \multirow{2}{*}{\textbf{Prompt}} &  \multicolumn{3}{c|}{\textbf{Irish}} &  \multicolumn{3}{c|}{\textbf{Welsh}} &  \multicolumn{3}{c|}{\textbf{Breton}}\\
       & &  \fontsize{8}{4}\selectfont\textbf{BERT-P}$\uparrow$ & \fontsize{8}{4}\selectfont\textbf{BERT-R}$\uparrow$ & \fontsize{8}{4}\selectfont\textbf{BERT-F1}$\uparrow$ &  \fontsize{8}{4}\selectfont\textbf{BERT-P}$\uparrow$ & \fontsize{8}{4}\selectfont\textbf{BERT-R}$\uparrow$ & \fontsize{8}{4}\selectfont\textbf{BERT-F1}$\uparrow$ &  \fontsize{8}{4}\selectfont\textbf{BERT-P}$\uparrow$ & \fontsize{8}{4}\selectfont\textbf{BERT-R}$\uparrow$ & \fontsize{8}{4}\selectfont\textbf{BERT-F1}$\uparrow$ \\
\hline
\multirow{8}{*}{\rotatebox{90}{\fontsize{8}{4}\selectfont GPT-3.5 (175B)}} &        ZS MI &            0.7574 &         0.7543 &     0.7555 &            0.7837 &         0.7796 &     0.7813 &            0.7768 &         0.7688 &     0.7722 \\
            &        FS IC &            0.7723 &         0.7661 &     0.7688 &            0.8057 &         0.7928 &     0.7989 & \textit{\textbf{0.7979}} & \textit{\textbf{0.7817}} & \textit{\textbf{0.7892}} \\
            &   ZS MI + GT &            0.8115 &         0.8035 &     0.8071 &            0.8255 &         0.8253 &     0.8251 & - & - & - \\
            &   FS IC + GT & \textit{\textbf{0.8149}} & \textit{\textbf{0.8044}} &  \textit{\textbf{0.8093}} & \textit{\textbf{0.8283}} & \textit{\textbf{0.8259}} & \textit{\textbf{0.8268}} & - & - & - \\
            &   ZS MI + AT &            0.8077 &         0.7973 &     0.8022 &            0.8217 &         0.8213 &     0.8212 &            0.7595 &         0.7384 &     0.7482 \\
            &   FS IC + AT &            0.8107 &         0.7984 &     0.8041 &            0.8253 &         0.8227 &     0.8237 &            0.7618 &         0.7379 &      0.749 \\
            & ZSMI+NLLB &            0.7998 &         0.7824 &     0.7906 &            0.8149 &         0.7979 &     0.8057 & - & - & - \\
            & FSIC+NLLB &            0.8025 &         0.7824 &     0.7919 &            0.8176 &         0.7977 &      0.807 & - & - & - \\
            \hline
\multirow{8}{*}{\rotatebox{90}{\fontsize{8}{4}\selectfont BLOOM (176B)}} &        ZS MI &            0.6485 &         0.6282 &     0.6365 &            0.6166 &         0.6265 &       0.62 &             0.598 &         0.6181 &     0.6057 \\
            &        FS IC &            0.6857 &         0.6757 &     0.6797 &            0.7173 &         0.6928 &     0.7035 &            0.7178 &          0.699 &     0.7071 \\
            &   ZS MI + GT &            0.7432 &         0.7479 &     0.7442 &            0.7533 &         0.7641 &     0.7572 &  - & - & - \\
            &   FS IC + GT & \textit{0.7829} & \textit{0.7758} & \textit{0.7786} & \textit{0.7933} & \textit{0.7921} & \textit{0.7918} & - & - & - \\
            &   ZS MI + AT &            0.7406 &         0.7435 &     0.7408 &            0.7514 &         0.7618 &     0.7552 &            0.7107 &          0.703 &     0.7054 \\
            &   FS IC + AT &            0.7758 &         0.7695 &     0.7718 &            0.7897 &         0.7893 &     0.7886 & \textit{0.7428} & \textit{0.7247} & \textit{0.7325} \\
            & ZSMI+NLLB &            0.6391 &         0.6241 &     0.6305 &            0.6497 &         0.6235 &     0.6353 & - & - & - \\
            & FSIC+NLLB &            0.6525 &         0.6324 &     0.6416 &            0.6642 &         0.6308 &     0.6463 & - & - & - \\
            \hline
\multirow{8}{*}{\rotatebox{90}{\fontsize{8}{4}\selectfont LLaMa2-chat (70B)}} &        ZS MI &            0.7051 &         0.6563 &     0.6781 &            0.7153 &         0.6742 &     0.6926 &            0.7214 &         0.6539 &     0.6843 \\
            &        FS IC &            0.7324 &         0.7278 &     0.7295 &            0.7272 &         0.7273 &     0.7265 &            0.7371 &         0.7101 &     0.7225 \\
            &   ZS MI + GT &            0.7909 &           0.79 &     0.7897 &            0.8025 &         0.8079 &     0.8043 & - & - & - \\
            &   FS IC + GT & \textit{0.8046} & \textit{0.8007} & \textit{0.8023} & \textit{0.8168} & \textit{0.8208} & \textit{0.8184} & - & - & - \\
            &   ZS MI + AT &             0.787 &         0.7847 &     0.7852 &             0.799 &         0.8054 &     0.8014 &            0.7461 &         0.7295 &     0.7368 \\
            &   FS IC + AT &               0.8 &         0.7949 &      0.797 &            0.8129 &         0.8176 &     0.8149 & \textit{0.7554} & \textit{0.737} & \textit{0.7453} \\
            & ZSMI+NLLB &            0.7834 &         0.7663 &     0.7739 &            0.7974 &          0.781 &     0.7881 & - & - & - \\
            & FSIC+NLLB &            0.7949 &         0.7789 &     0.7862 &            0.8088 &         0.7931 &     0.8002 & - & - & - \\
            \hline
\multirow{8}{*}{\rotatebox{90}{\fontsize{8}{4}\selectfont Falcon-chat (180B)}} &        ZS MI &            0.6961 &         0.6833 &     0.6885 &            0.7004 &         0.6854 &     0.6914 &            0.7232 &         0.6839 &     0.7013 \\
            &        FS IC &            0.7384 &         0.7397 &     0.7385 &              0.77 &           0.75 &     0.7589 &            0.7412 &         0.7119 &     0.7253 \\
            &   ZS MI + GT &            0.7656 &         0.7792 &     0.7712 &            0.7758 &         0.7967 &     0.7849 & - & - & - \\
            &   FS IC + GT & \textit{0.8029} & \textit{0.8003} & \textit{0.8012} & \textit{0.8155} & \textit{0.8221} & \textit{0.8183} & - & - & - \\
            &   ZS MI + AT &            0.7623 &         0.7743 &     0.7672 &            0.7743 &         0.7944 &      0.783 &            0.7307 &          0.726 &     0.7273 \\
            &   FS IC + AT &            0.7983 &          0.795 &     0.7962 &            0.8135 &         0.8197 &     0.8162 & \textit{0.7566} & \textit{0.7387} & \textit{0.7468} \\
            & ZSMI+NLLB &            0.7616 &         0.7594 &     0.7594 &            0.7748 &          0.774 &     0.7732 & - & - & - \\
            & FSIC+NLLB &            0.7933 &         0.7784 &     0.7852 &            0.8094 &         0.7946 &     0.8012 & - & - & - \\
\hline
\end{tabular}
    \caption{BERTScore results for \textbf{Irish}, \textbf{Welsh} and \textbf{Breton}. Maltese is not available in BERTScore. Highest score in each column for each language in bold, highest score for each model in italics. Number of parameters in brackets in column~1. ZS MI=Zero-Shot Minimal Instruction, FS IC=Few-Shot In Context, GT=Google Translate, AT=Alibaba Translate, NLLB=No Language Left Behind system.}
    \label{tab:bert-res}
\end{table*}

\begin{table}[h!tb]
    \centering
    \small
    \setlength\tabcolsep{4.5pt} 
    \renewcommand{\arraystretch}{1.15}
    \begin{tabular}{|ll|ccc|}
\hline
             \multirow{2}{*}{\textbf{Model}} & \multirow{2}{*}{\textbf{Prompt}} & \multicolumn{3}{c|}{\textbf{BERT}} \\
              &  & \textbf{P}$\uparrow$ & \textbf{R}$\uparrow$ & \textbf{F1}$\uparrow$ \\
\hline
    GPT-3.5 & ZS MI &         0.9555 &      0.9568 &  0.9555 \\
    (175B) & FS IC &         \textit{\textbf{0.9588}} &      \textit{\textbf{0.9582}} &   \textit{\textbf{0.958}} \\
    \hline
      BLOOM & ZS MI &         0.9092 &      0.9234 &  0.9151 \\
      (176B) & FS IC &          \textit{0.938} &       \textit{0.937} &  \textit{0.9368} \\
      \hline
 LLaMa2-chat & ZS MI &         0.9449 &      0.9465 &  0.9449 \\
 (70B) & FS IC &         \textit{0.9522} &      \textit{0.9535} &  \textit{0.9523} \\
 \hline
Falcon-chat & ZS MI &         0.9276 &      0.9379 &  0.9319 \\
(180B) & FS IC &         \textit{0.9532} &      \textit{0.9543} &  \textit{0.9531} \\
\hline
\end{tabular}
    \caption{BERTScore results in \textbf{English}. Best score per column in bold, best score per model in italics. Number of model parameters in brackets. ZS MI=Zero-Shot Minimal Instruction, FS IC=Few-Shot In Context.}
    \label{tab:res-en-bert}
\end{table}

\begin{table}[h!tb]
    \centering
    \small
    \setlength\tabcolsep{2pt} 
    \renewcommand{\arraystretch}{1.15}
    \begin{tabular}{|l|l|c|c|c|c|c|}
\hline
            \multirow{2}{*}{\textbf{M}} & \multirow{2}{*}{\textbf{Prompt}} & \multicolumn{5}{c|}{\textbf{COMET} $\uparrow$}\\ \cline{3-7}
              &  & \textbf{Irish} & \textbf{Welsh} & \textbf{Maltese} & \textbf{Breton} & \textbf{English} \\
\hline
\multirow{8}{*}{\rotatebox{90}{\fontsize{8}{4}\selectfont GPT-3.5 (175B)}} &        ZS MI &   0.6606 &   0.7301 &   0.6378 &   0.6772 &   0.8261 \\
                   &        FS IC &   0.6994 &   0.7521 &   0.6425 &   \textit{\textbf{0.6962}} &   \textit{\textbf{0.8306}} \\
                   &   ZS MI + GT &   0.7387 &   0.7918 &    0.676 &        - &        - \\
                   &   FS IC + GT &   \textit{0.7431} &   \textit{\textbf{0.7939}} &   \textit{\textbf{0.6739}} &        - &        - \\
                   &   ZS MI + AT &   0.7205 &   0.7776 &   0.6584 &   0.5698 &        - \\
                   &   FS IC + AT &   0.7279 &   0.7796 &   0.6557 &   0.5711 &        - \\
                   & ZSMI+NLLB &   0.7155 &   0.7513 &   0.6583 &        - &        - \\
                   & FSIC+NLLB &    0.715 &   0.7542 &   0.6584 &        - &        - \\
                   \hline
\multirow{8}{*}{\rotatebox{90}{\fontsize{8}{4}\selectfont BLOOM (176B)}} &        ZS MI &   0.4525 &   0.4152 &   0.4426 &    0.428 &   0.7186 \\
                   &        FS IC &   0.4523 &   0.4837 &   0.5401 &   0.5242 &   \textit{0.7799} \\
                   &   ZS MI + GT &   0.6569 &   0.7015 &   0.6125 &        - &        - \\
                   &   FS IC + GT &   \textit{0.7063} &   \textit{0.7512} &   \textit{0.6426} &        - &        - \\
                   &   ZS MI + AT &   0.6459 &   0.6865 &    0.602 &    0.522 &        - \\
                   &   FS IC + AT &   0.6884 &   0.7386 &   0.6274 &   \textit{0.5596} &        - \\
                   & ZSMI+NLLB &   0.6327 &   0.6686 &   0.6027 &        - &        - \\
                   & FSIC+NLLB &   0.6812 &   0.7191 &   0.6289 &        - &        - \\
                   \hline
\multirow{8}{*}{\rotatebox{90}{\fontsize{8}{4}\selectfont LLaMa2-chat (70B)}} &        ZS MI &   0.4761 &   0.4775 &   0.5403 &    0.416 &   0.7962 \\
                   &        FS IC &   0.5959 &    0.541 &   0.5923 &   0.4866 &   \textit{0.8211} \\
                   &   ZS MI + GT &   0.7204 &   0.7662 &    0.655 &        - &        - \\
                   &   FS IC + GT &   \textit{0.7381} &   \textit{0.7856} &   \textit{0.6689} &        - &        - \\
                   &   ZS MI + AT &   0.7005 &   0.7546 &   0.6386 &   0.5583 &        - \\
                   &   FS IC + AT &   0.7185 &   0.7722 &   0.6492 &   \textit{0.5696} &        - \\
                   & ZSMI+NLLB &    0.688 &   0.7303 &   0.6369 &        - &        - \\
                   & FSIC+NLLB &   0.7092 &   0.7495 &    0.648 &        - &        - \\
                   \hline
\multirow{8}{*}{\rotatebox{90}{\fontsize{8}{4}\selectfont Falcon-chat (180B)}} &        ZS MI &   0.5393 &   0.5437 &   0.5331 &   0.4854 &    0.765 \\
                   &        FS IC &   0.6182 &   0.6566 &    0.599 &   0.5599 &   \textit{0.8229} \\
                   &   ZS MI + GT &   0.7063 &   0.7487 &   0.6363 &        - &        - \\
                   &   FS IC + GT &   \textit{\textbf{0.7457}} &   \textit{0.7922} &   \textit{0.6709} &        - &        - \\
                   &   ZS MI + AT &   0.6866 &   0.7371 &   0.6227 &   0.5519 &        - \\
                   &   FS IC + AT &   0.7257 &   0.7817 &   0.6534 &   \textit{0.5731} &        - \\
                   & ZSMI+NLLB &   0.6809 &   0.7186 &    0.622 &        - &        - \\
                   & FSIC+NLLB &   0.7154 &   0.7546 &   0.6498 &        - &        - \\
\hline
\end{tabular}
    \caption{COMET results for \textbf{Irish}, \textbf{Welsh}, \textbf{Maltese}, \textbf{Breton}, and \textbf{English}. COMET scores are between 0 and 1. Highest score in each column for each language in bold, highest score for each model in italics. Number of parameters in brackets in column~1. ZS MI=Zero-Shot Minimal Instruction, FS IC=Few-Shot In Context, GT=Google Translate, AT=Alibaba Translate, NLLB=No Language Left Behind system.}
    \label{tab:res-comet}
\end{table}

\section{Prompts}
\label{sec:appendix-prompts}
We provide the prompts we used to execute all our experiments. In Table\ \ref{tab:temp-zero-shot} Zero-Shot Minimal Instruction prompt is shown, while in Table\ \ref{tab:temp-few-shot} Few-Shot In Context prompt is shown with the examples used for each language tested.

\begin{table*}[h!]
\centering\small
\renewcommand{\arraystretch}{1.15}
\begin{tabular}{|>{\raggedright\arraybackslash}p{0.15\textwidth}|>{\raggedright\arraybackslash}p{0.75\textwidth}|}
\hline
\multicolumn{2}{|c|}{\textbf{Zero-Shot Minimal Instruction}} \\
\hline
\textbf{Template:} & Write the following triples as fluent {English | Irish | Welsh | Maltese | Breton} text.\bigbreak Triples: """\par \{set of triples in the format \textit{subject predicate object} and each triple in a new line\}\par """\bigbreak Text: [MODEL] \\
\hline
\end{tabular}
\caption{\label{tab:temp-zero-shot}
Template of the Zero-Shot Minimal Instruction prompt.}
\end{table*}

\begin{table*}[h!]
\centering\small
\renewcommand{\arraystretch}{1.15}
\begin{tabular}{|>{\raggedright\arraybackslash}p{0.15\textwidth}|>{\raggedright\arraybackslash}p{0.75\textwidth}|}
\hline
\multicolumn{2}{|c|}{\textbf{Few-Shot In Context}} \\
\hline
\textbf{Template:} & Write the following triples as fluent {English | Irish | Welsh | Maltese | Breton} text.\bigbreak Triple 1: """\par \{set of triples in the format \textit{subject predicate object} and each triple in a new line\}\par """\par Text 1: \{verbalisation of Triple 1\}\par \#\#\par Triple 2: """\par \{set of triples in the format \textit{subject predicate object} and each triple in a new line\}\par """\par Text 2: \{verbalisation of Triple 2\}\par \#\#\par Triple 3: """\par \{set of triples in the format \textit{subject predicate object} and each triple in a new line\}\par """\par Text 3: [MODEL] \\
\hline \hline
\textbf{English, Irish, and Breton Triples:} & Triple 1: Adolfo\_Suárez\_Madrid–Barajas\_Airport runwayName "14R/32L" \bigbreak Triple 2: American\_Journal\_of\_Mathematics abbreviation "Am. J. Math." \par American\_Journal\_of\_Mathematics firstPublicationYear 1878 \par American\_Journal\_of\_Mathematics issnNumber "1080-6377" \\
\hline
\textbf{English texts:} & Text 1: 14R/32L is the runway name of Adolfo Suárez Madrid-Barajas Airport. \par Text 2: The American Journal of Mathematics was first published in 1878 and is also known by the abbreviated title of Am. J. Math. It has an ISSN number of 1080-6377. \\
\hline
\textbf{Irish texts:} & Text 1: 14R/32L is ainm do rúidbhealach Aerfort Adolfo Suárez Madrid-Barajas \par Text 2: Foilsíodh an American Journal of Mathematics don chéad uair in 1878 agus aithnítear leis an ainm giorraithe Am. J. Math. chomh maith é. Tá an uimhir ISSN 1080-6377 aige. \\
\hline
\textbf{Breton texts:} & Text 1: Anv leurenn bradañ aerborzh Adolfo Suárez Madrid-Barajas zo 14L/32R. \par Text 2: Finland zo bro ar Finniz hag hini ar skorndorrer Aleksey Chirikov bet savet e chanter-bigi Arctech en Helsinki. \\
\hline \hline  
\textbf{Maltese and Welsh Triples:} & Triple 1: Albennie\_Jones birthPlace Errata,\_Mississippi \bigbreak Triple 2: GMA\_New\_Media industry Entertainment\par GMA\_New\_Media type Media\_company\par GMA\_New\_Media product World\_Wide\_Web \\
\hline
\textbf{Maltese texts:} & Text 1: Albennie Jones twieldet f'Errata Mississippi. \par Text 2: GMA New Media hija kumpanija tal-midja tal-industrija tad-divertiment li toffri servizzi li jikkonċernaw il-World Wide Web. \\
\hline
\textbf{Welsh texts:} & Text 1: Ganed Albennie Jones yn Errata, Mississippi. \par Text 2: Mae GMA New Media yn gwmni cyfryngau yn y diwydiant adloniant sy'n cynnig gwasanaethau sy'n ymwneud â'r We Fyd Eang. \\
\hline
\end{tabular}
\caption{\label{tab:temp-few-shot}
Few-Shot In Context prompt. \textbf{Top} Template of the prompt. \textbf{Center} Examples' triple set and texts in English, Irish, and Breton. \textbf{Bottom} Examples' triple set and texts in Maltese and Welsh.}
\end{table*}

\section{Human evaluation setup}
For our human evaluation of English systems, we considered the human-authored references, GPT-3.5 Zero-Shot Minimal Instruction prompt, GPT-3.5 Few-Shot In Context prompt, and the best WebNLG2020 system (Amazon AI). For each system, we annotated 100 samples recruiting 4 annotators, who are non-author members 
of the research group 
plus one close collaborator.

We followed the same annotation guidelines provided by \citet{cripwell-etal-2023}.

In Figure\ \ref{fig:hum-eval}, the screenshot of the human evaluation interface given to the annotators is shown.

\begin{figure*}[h!tb]
\centering
    \includegraphics[width=\textwidth]{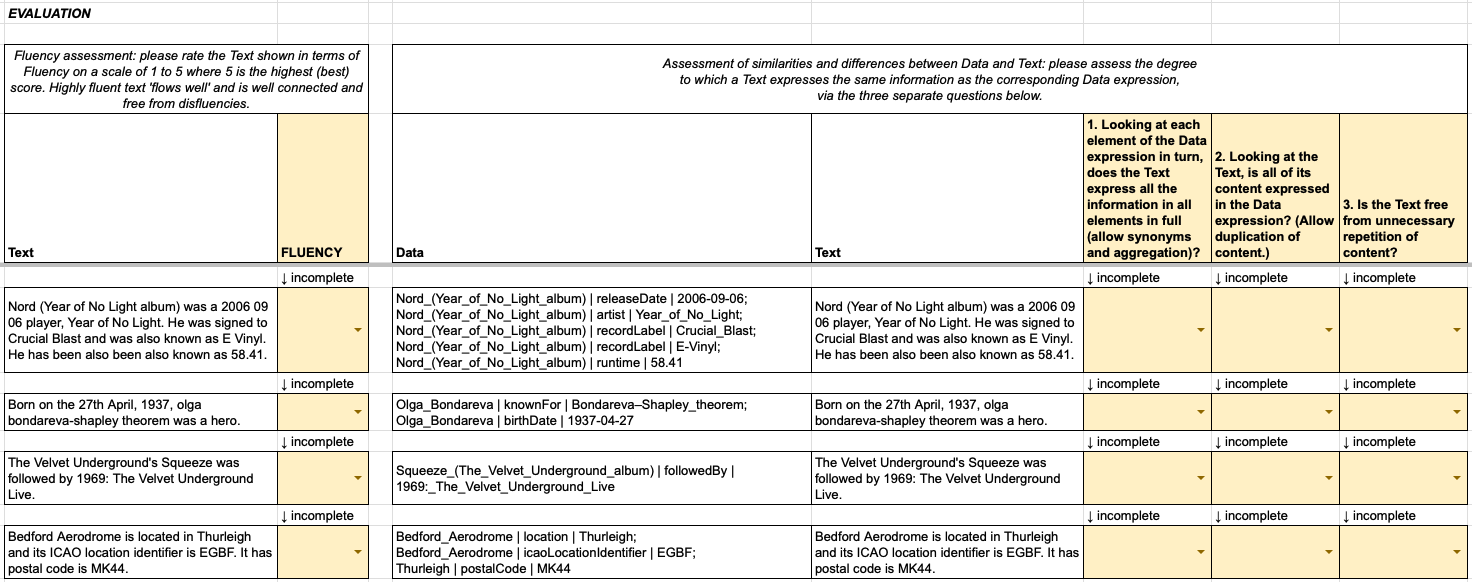}
    \caption{Screenshot of the human evaluation interface.}
\end{figure*}\label{fig:hum-eval}

\section{Scientific artifacts and licensing}
In this work, we used the following scientific artifacts. BLOOM is licensed under The BigScience RAIL License. LLaMa2 is licensed under a commercial license \footnote{\url{https://ai.meta.com/resources/models-and-libraries/llama-downloads/}}. GPT-3.5 is licensed under a commercial license \footnote{\url{https://openai.com/policies/terms-of-use}}. Falcon is licensed under the FALCON 180B TII LICENSE VERSION 1.0 \footnote{\url{https://huggingface.co/tiiuae/falcon-180B-chat/blob/main/ACCEPTABLE_USE_POLICY.txt}}. NLLB is licensed under CC-BY-NC-4.0 \footnote{\url{https://huggingface.co/facebook/nllb-200-1.3B/blob/main/README.md}}. The usage of the listed artifacts is consistent with their licenses.

\end{document}